\documentclass[letterpaper]{article} 
\usepackage{aaai2026}  
\usepackage{times}  
\usepackage{helvet}  
\usepackage{courier}  
\usepackage[hyphens]{url}  
\usepackage{graphicx} 
\usepackage{natbib}  
\usepackage{caption} 
\usepackage{algorithm}
\usepackage{algorithmic}

\usepackage{newfloat}
\usepackage{listings}
\usepackage{amsmath}
\usepackage{amssymb}  
\usepackage{amsfonts} 
\usepackage{multirow}
\usepackage{booktabs} %
\usepackage{array}
\usepackage{xcolor}
\usepackage{tabularx}
\usepackage{makecell}
\usepackage{newfloat}
\usepackage{listings}
\DeclareCaptionStyle{ruled}{labelfont=normalfont,labelsep=colon,strut=off} 
\floatstyle{ruled}
\newfloat{listing}{tb}{lst}{}
\floatname{listing}{Listing}
\title{\textsf{\textcolor{orange}{M}\textcolor{orange}{P}\textcolor{orange}{1}}: \textsf{\textcolor{orange}{M}}eanFlow Tames \textsf{\textcolor{orange}{P}}olicy Learning in \textsf{\textcolor{orange}{1}}-step for Robotic Manipulation}
\author{
    Juyi Sheng\textsuperscript{1}\equalcontrib,
    Ziyi Wang\textsuperscript{1}\equalcontrib,
    Peiming Li\textsuperscript{1},
    Mengyuan Liu\textsuperscript{1,}\thanks{Corresponding authors.}
}
\affiliations{
    \textsuperscript{\rm 1}
    State Key Laboratory of General Artificial Intelligence, \\Peking
    University, Shenzhen Graduate School, Shenzhen, China
}

\usepackage{bibentry}

\begin{document}

\maketitle

\begin{abstract}

In robot manipulation, robot learning has become a prevailing approach. However, generative models within this field face a fundamental trade-off between the slow, iterative sampling of diffusion models and the architectural constraints of faster Flow-based methods, which often rely on explicit consistency losses. To address these limitations, we introduce MP1, which pairs 3D point-cloud inputs with the MeanFlow paradigm to generate action trajectories in one network function evaluation (1-NFE). By directly learning the interval-averaged velocity via the ``MeanFlow Identity", our policy avoids any additional consistency constraints. This formulation eliminates numerical ODE-solver errors during inference, yielding more precise trajectories. MP1 further incorporates CFG for improved trajectory controllability while retaining 1-NFE inference without reintroducing structural constraints. Because subtle scene-context variations are critical for robot learning, especially in few-shot learning, we introduce a lightweight \textit{Dispersive Loss} that repels state embeddings during training, boosting generalization without slowing inference. We validate our method on the Adroit and Meta-World benchmarks, as well as in real-world scenarios. Experimental results show MP1 achieves superior average task success rates, outperforming DP3 by 10.2\% and FlowPolicy by 7.3\%. Its average inference time is only 6.8 ms—19× faster than DP3 and nearly 2× faster than FlowPolicy. Our project page is available at \url{https://mp1-2254.github.io/}, and the code can be accessed at \url{https://github.com/LogSSim/MP1}.

\end{abstract}

\section{Introduction}

Robot manipulation refers to the process by which robots develop the ability to perform physical tasks, such as grasping, moving, and manipulating objects. A fundamental approach in this domain is robot learning, which enables robots to generate actions based on visual inputs (such as images and point clouds) or textual descriptions. Recent advancements in robot learning have been driven by methods like Transformers \cite{transformer}, Diffusion Models \cite{dp}, and Flow Matching \cite{fm}. These methods have enhanced the ability of robots to understand and execute complex actions in response to multimodal inputs \cite{dp1,dp2}, facilitating more effective and versatile robot learning.

\begin{figure}[t]
    \centering
    \includegraphics[width=0.9\linewidth]{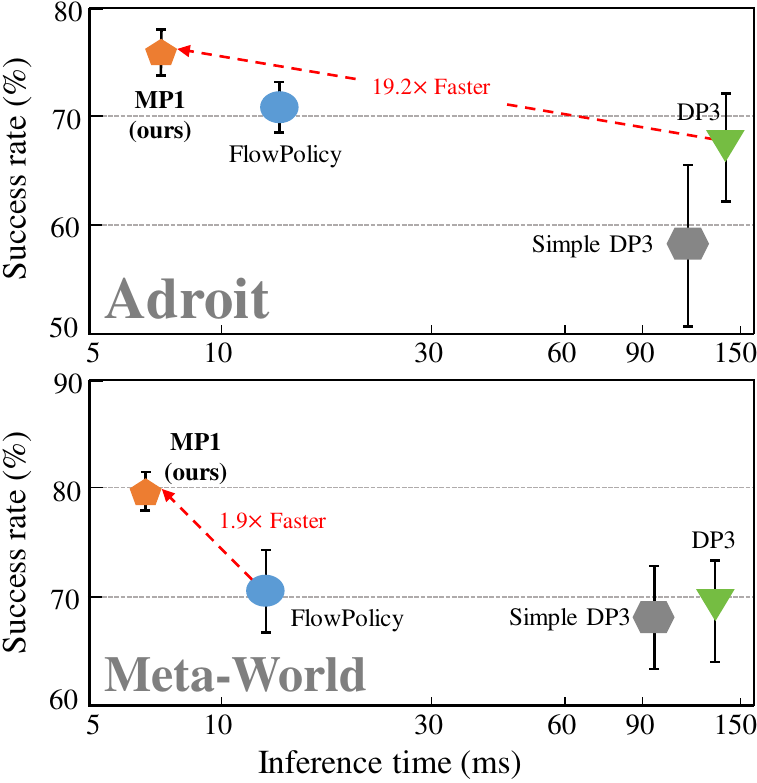}
    \caption{The proposed method outperforms SOTA methods (DP3 \cite{dp3} and FlowPolicy \cite{flowpolicy}) on the Adroit and Meta-World tasks, showing superior inference time and success rate, as demonstrated by the MP1 on the comparison plot.}
    \label{scatter}
\end{figure}

Among them, Diffusion Model-based methods, such as the diffusion policy \cite{dp1, dp2}, effectively handle multimodal action distributions by representing robot action predictions as probability distributions. Additionally, the DP3 \cite{dp3} introduces 3D point cloud features combined with the diffusion policy, improving the success rate of tasks and reducing the amount of training data. Through a step-by-step denoising process, Diffusion Models can capture multiple possible action choices, thereby increasing the flexibility and accuracy of action generation. However, a notable drawback of Diffusion Models is their relatively long inference time. Since action generation requires multiple time steps to denoise, the inference process can be time-consuming, which may become a bottleneck in applications that demand real-time performance.

In recent years, Flow-based methods have been proposed to overcome this problem, such as AdaFlow \cite{adaflow}, FlowPolicy \cite{flowpolicy}, which aims to reduce sampling steps and achieve more efficient single-step inference. These methods greatly accelerate inference by optimizing the generative process, but they typically require additional consistency constraints on the model’s outputs to ensure valid trajectories. In contrast, the recently introduced MeanFlow \cite{meanflow} paradigm from image generation avoids any explicit consistency loss by learning a mean velocity field. This innovation simplifies the generation process and achieves genuine one-step sampling, markedly improving real-time performance. For example, MeanFlow can reduce inference latency to around 6.8\,ms, far better than the 10–20 steps required by diffusion strategies.

We present the first adaptation of the MeanFlow \cite{meanflow} paradigm to robot learning, termed MP1. Conditioned on 3D point-cloud observations, MP1 learns the interval-averaged velocity and bypasses the need to integrate instantaneous velocities. This design eliminates ODE-solver error and yields genuinely one-step \mbox{(1-NFE)} trajectory generation with smooth, dynamically consistent actions. However, a purely regression-based objective fails to impose explicit regularization on the policy’s internal feature space \cite{Disperse}. This representational ambiguity is particularly detrimental in robotics, where subtle variations in scene context are critical, and it undermines generalization in few-shot learning.

To counter this, we add Dispersive Loss, spreading out the latent embeddings of distinct states. Acting as a contrastive-style regularizer without positive pairs, it sharpens state discrimination while the original regression term still aligns each state to its expert trajectory. With only ten expert demonstrations, this combination already closes most performance gaps (Fig.~\ref{figure:ablation}), underscoring the method’s few-shot generalization. Because Dispersive Loss is computed once per forward pass and vanishes at inference, MP1 preserves its hallmark 1-NFE speed.
We conduct experiments on the Adroit and Meta-World simulation datasets and real-world tasks. MP1 is capable of one-step inference and, compared to state-of-the-art (SOTA) methods, improves the average success rate by 7.3\% (Tab. \ref{success}) while also achieving superior inference speed (Tab. \ref{inference}). Our contributions are as follows:
\begin{itemize}
    \item We introduce MP1, the first MeanFlow-based robot-learning framework. Conditioning on 3D point-cloud features, it learns effectively from a handful of demonstrations, yet delivers one-step sampling with SOTA success rates and millisecond-level inference latency.
    \item We incorporate a lightweight Dispersive Loss that regularizes latent features without affecting runtime, boosting few-shot generalization.
    \item Extensive experiments on Adroit and Meta-World simulation benchmarks, as well as real-world tasks, show that MP1 surpasses Diffusion- and Flow-based baselines in both success rate and speed.
\end{itemize}

\begin{figure*}[ht]
    \centering
    \includegraphics[width=0.95\linewidth]{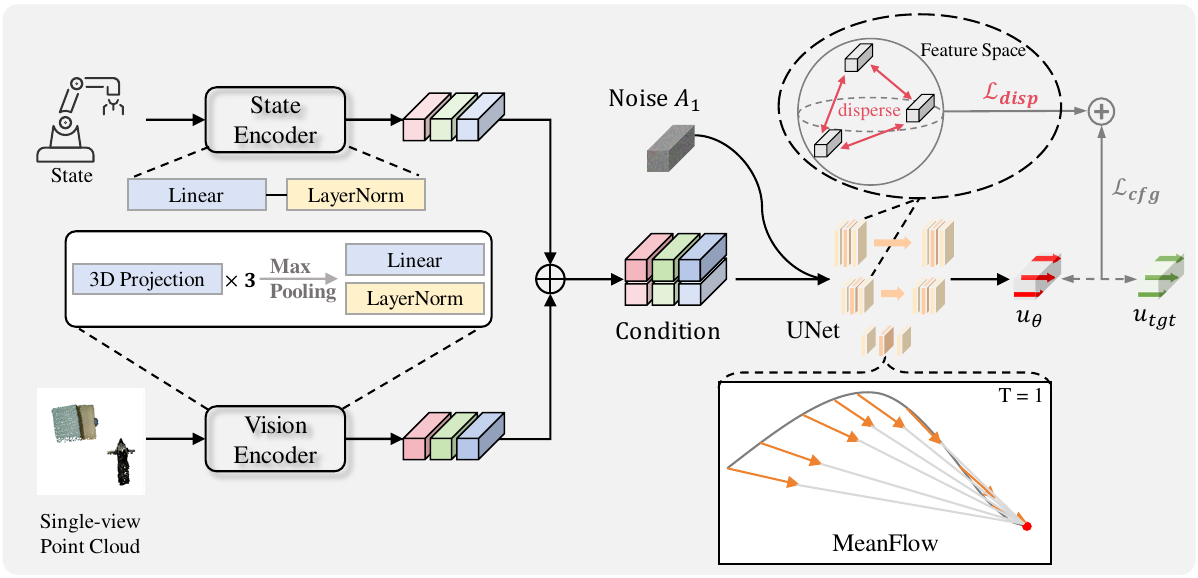}
    \caption{Overview of MP1. The MP1 takes the historical observation point cloud and the robot's state as inputs. These inputs are processed through a visual encoder and a state encoder, respectively, and then serve as conditional inputs to the UNet-integrated MeanFlow. After passing through the MeanFlow, the model computes regression loss ($\mathcal{L}_{cfg}$) between the mean velocity generated from the initial noise and the target velocity. This $\mathcal{L}_{cfg}$ is combined with a Dispersive Loss ($\mathcal{L}_{disp}$) imposed on the UNet’s hidden states to jointly optimize the network parameters. }
    \label{pipeline}
    \label{}
\end{figure*}

\section{Related work}
\subsection{2D Input Robot Learning}

Most methods that utilize 2D visual input predict robot actions based on images. For example, BC-Z \cite{bcz} aligned 2D visual-language features, allowing the robot to generalize to new target tasks. ALOHA \cite{aloha} employed an ACT network to model the relationship between 2D vision and robotic actions. Similarly, other approaches, such as DP \cite{dp1, dp2}, HPT \cite{hpt}, also leveraged 2D inputs for robot learning policies. However, 2D inputs often lack depth information, which limits the accuracy in completing tasks.

\subsection{3D Input Robot Learning}

To overcome the limitations of 2D inputs, 3D inputs have gained prominence. Approaches such as PerACT \cite{peract} and ACT3D \cite{act3d} utilized voxel data. However, due to the high computational cost associated with voxel data, point clouds have become the dominant form of 3D input. Methods like RVT \cite{rvt}, RVT2 \cite{rvt2}, and DP3 \cite{dp3} leveraged point clouds, and their introduction has been shown to enhance task success rates.
\subsection{Diffusion-Based Robot Learning}

Diffusion models have recently made significant progress in image and action generation. This model simulates the diffusion process of data by gradually adding noise and then denoising to generate data. In robot learning, diffusion models are used to generate continuous control actions and strategies to tackle complex tasks.
Initially, diffusion policy \cite{dp1, dp2} addressed the issues of multimodal action generation and action consistency. Later, DP3 \cite{dp3} introduced point cloud information, greatly improving task success rates. HDP \cite{hdp} established spatial relationships between task points in point clouds using diffusion. RDT \cite{rdt} built a large-scale robot dual-arm model based on diffusion. Additionally, models like EquiDiff \cite{equ} and Instant Policy \cite{instant} are also based on diffusion. However, diffusion still faces challenges related to inference time.

\subsection{Flow-Based Robot Learning}

Flow matching is a method for training continuous normalized flows by learning vector fields associated with probabilistic paths. Recent research has shown that Flow matching performs well in image generation \cite{flowmatching1, flowmatching2, flowmatching3}, and Consistency Models \cite{consistencymodels1, consistencymodels2} further improve the sampling efficiency of Flow methods, enabling one-step sampling. Consequently, AdaFlow \cite{adaflow} was introduced as a robot learning framework using flow‐based generative modeling to represent the policy via state‐conditioned ordinary differential equations. Building on this approach, FlowPolicy \cite{flowpolicy} adapted it to robot learning, achieving one‐step sampling by enforcing input consistency constraints.

3D point clouds are particularly well-suited for robot learning \cite{act3d, dp3}, which is why this paper utilizes them as input. Methods based on diffusion and flow focus on understanding the environment and generating actions. While diffusion-based methods effectively address these challenges, they introduce significant delays in inference time due to the need for multiple sampling steps. On the other hand, Flowpolicy achieves one-step sampling by applying consistency constraints to the input, but it relies on certain assumptions. The MP1 method proposed in this paper is derived from the definitions of average and instantaneous velocities. This approach eliminates the need for assumption-based constraints, enabling efficient one-step inference and the generation of high-quality robot actions.

\section{Method}

Our goal is to develop a robot learning policy that is both highly efficient and robust. Efficiency, particularly single-step (1-NFE) inference, is crucial for real-time applications, while robustness is essential for generalizing from limited demonstrations and handling subtle environmental variations. To address these challenges, we propose the MP1 (in Fig. 2). Unlike Diffusion-based methods, our approach does not require multi-step denoising; distinct from existing Flow-based approaches, the MP1 does not rely on ODE solvers, consistency constraints, and the integration of interval-wise instantaneous velocities. Furthermore, by encouraging the latent embeddings of different input states to disperse, we improve the model’s generalization abilities and task success rate, all without sacrificing inference speed.

\begin{figure*}[!htbp]
    \centering
    \includegraphics[width=1.0\linewidth]{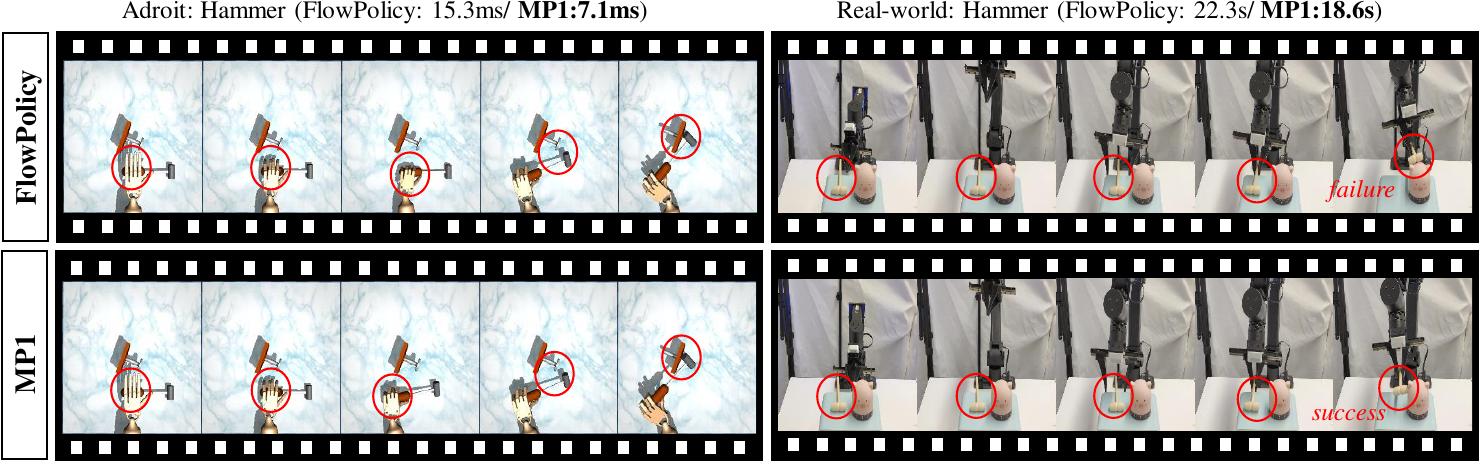}
    \caption{Qualitative comparison of the proposed MP1 and the previous SOTA method (FlowPolicy \cite{flowpolicy}) on Adroit Hammer and real-world Hammer tasks. Our method is faster, with 7.1ms in the simulated hammer and 18.6s in the real-world scenario. Moreover, our method successfully completes the real-world hammer task, whereas FlowPolicy fails.}
    \label{fig:enter-label}
\end{figure*}

\subsection{MP1: One-Step Trajectory Generation}
In the context of robot learning, the policy's task is to map a sequence of observations, including 3D point clouds $\mathbf{P}$ and robotic states $\mathbf{S}$, to a future action trajectory $\mathbf{A}$. We first process these inputs using encoders to obtain conditional features. The point cloud $\mathbf{P} \in \mathbb{R}^{n_{o} \times np \times 3}$ is passed through 3D Projection to produce a visual feature vector $\mathbf{f}_v$, while the robot state $\mathbf{S} \in \mathbb{R}^{n_{o} \times s_{d}}$ is encoded into a state feature $\mathbf{f}_s$. These are combined into a single conditional vector $\mathbf{c} = (\mathbf{f}_v, \mathbf{f}_s)$ that guides the action generation process.

To achieve single-step generation, we model the policy as a conditional MeanFlow. Unlike standard Flow Matching (FM), which learns an instantaneous velocity field $v(z_t, t)$ and requires solving an ODE for sampling, MeanFlow learns the average velocity field $u(z_t, r, t)$ over an interval $[r, t]$. The average velocity is defined as the total displacement divided by the time duration:
\begin{equation}
u(z_{t},r,t)\triangleq\frac{1}{t-r}\int_{r}^{t}v(z_{\tau},\tau)d\tau
\end{equation}
This formulation is key to bypassing iterative integration. Training a network $u_{\theta}$ to model this field directly from its integral definition is intractable. Instead, MeanFlow leverages the ``MeanFlow Identity", a local differential equation derived by differentiating the integral definition to $t$:
\begin{equation}\label{equ}
u(z_{t},r,t)=v(z_{t},t)-(t-r)\frac{d}{dt}u(z_{t},r,t)
\end{equation}
where the total derivative $\frac{d}{dt}u(z_{t},r,t)$ expands to $v(z_{t},t)\partial_{z}u+\partial_{t}u$. This allows us to train $u_{\theta}$ with a simple regression objective that enforces this identity:
\begin{equation}
\mathcal{L}(\theta)=\mathbb{E}_{t,r,x,\epsilon}||u_{\theta}(z_{t},r,t)-sg(u_{tgt})||_{2}^{2}
\end{equation}
The target $u_{tgt}$ is constructed using the known instantaneous velocity $v_t$ of the probability path and the network's own estimate of the total derivative, with a stop-gradient $sg(\cdot)$ to ensure stability:
\begin{equation}
u_{tgt}=v_{t}-(t-r)(v_{t}\partial_{z}u_{\theta}+\partial_{t}u_{\theta})
\end{equation}
For our MP1, we adapt this framework to generate action trajectories $\mathbf{A}$. The network $u_{\theta}(\mathbf{A}_t, r, t | \mathbf{c})$ is trained to predict the average velocity that transforms a noise vector $\mathbf{A}_1 \sim \mathcal{N}(0,I)$ into an expert action trajectory $\mathbf{A}_0$. To improve control, we integrate Classifier-Free Guidance (CFG) by training the network to model a CFG-aware average velocity, $u_{\theta}^{cfg}$. This is achieved with a modified regression loss based on a guided instantaneous velocity field:
\begin{equation}
\mathcal{L}_{cfg}(\theta) = \mathbb{E}||u_{\theta}^{cfg}(\mathbf{A}_t, r, t|\mathbf{c}) - sg(u_{tgt})||_{2}^{2}
\label{eq:cfg_loss}
\end{equation}
where the target $u_{tgt}$ is now computed using a guided velocity $\tilde{v}_{t} \triangleq \omega v_{t}(\mathbf{A}_t|\mathbf{A}_0, \mathbf{c}) + (1-\omega)u_{\theta}^{cfg}(\mathbf{A}_t, t, t| \emptyset)$, blending the conditional and unconditional predictions.

\subsection{Enhancing Representational generalization with Dispersive Loss}
While the MeanFlow objective $\mathcal{L}_{cfg}$ excels at learning the temporal dynamics required to produce accurate output trajectories, it provides only an indirect signal for learning the complex mapping from high-dimensional conditional inputs $\mathbf{c}$ to actions. This can lead to a form of ``feature collapse", where the policy network maps distinct environmental states that demand fundamentally different actions to nearly identical points in its latent space. Such ambiguity is particularly detrimental in robot learning, where subtle differences in object pose or scene configuration are critical for success, especially in few-shot learning regimes.

To directly address this, we incorporate Dispersive Loss, a principled and self-contained regularizer that operates on the model's internal representations. The core idea is to encourage the latent embeddings of different input samples within a training batch to disperse, thereby enforcing a more discriminative feature space. Dispersive Loss functions as a ``contrastive loss without positive pairs"; the repulsive force is supplied by the loss itself, while the alignment of a state to its correct action is handled implicitly by the primary $\mathcal{L}_{cfg}$ regression objective.
Let $\mathbf{z}_{\mathbf{A}} = f_{policy}(\mathbf{A}_t, t, \mathbf{c})$ be the intermediate representation from a chosen layer within our policy network. The Dispersive Loss is defined as:
\begin{equation} \label{eq:disp_loss}
\mathcal{L}_{Disp}(\theta) = \log \mathbb{E}_{i,j \in \mathcal{B}} \left[ \exp\left(-\frac{||\mathbf{z}_{\mathbf{A},i} - \mathbf{z}_{\mathbf{A},j}||_2^2}{\tau}\right) \right],
\end{equation}
where $\mathcal{B}$ is a mini-batch of training samples, $\mathbf{z}_{\mathbf{A},i}$ and $\mathbf{z}_{\mathbf{A},j}$ are the intermediate representations for two samples in the batch, and $\tau$ is a temperature hyperparameter. Specifically, we adopt the InfoNCE-based variant of Dispersive Loss (using $\ell_2$ distance, with temperature $\tau=1$) and apply it to the output features of each down-sampling block in the U-Net backbone of $u_{\theta}$.

\subsection{Training Objective and Inference}
Our final training objective synergistically combines the trajectory generation and representation regularization goals:
\begin{equation} \label{eq:total_loss}
\mathcal{L}_{total}(\theta) = \mathcal{L}_{cfg}(\theta) + \lambda \mathcal{L}_{Disp}(\theta)
\end{equation}
Here, $\mathcal{L}_{cfg}$ (Eq. \ref{eq:cfg_loss}) ensures the policy generates dynamically correct action trajectories, while $\mathcal{L}_{Disp}$ (Eq. \ref{eq:disp_loss}) promotes a well-structured and discriminative latent space to improve generalization and robustness. When $\lambda$ is set to 0.5, it balances the contribution of the two loss terms.

\begin{table*}[t]
    \centering
    \resizebox{1.0\linewidth}{!}{
    \begin{tabular}{
        l|
        >{\centering\arraybackslash}p{1.8cm}|
        >{\centering\arraybackslash}p{1.0cm}|
        >{\centering\arraybackslash}p{1.3cm}
        >{\centering\arraybackslash}p{1.3cm}
        >{\centering\arraybackslash}p{1.3cm}|
        >{\centering\arraybackslash}p{1.7cm}
        >{\centering\arraybackslash}p{1.7cm}
        >{\centering\arraybackslash}p{1.7cm}
        >{\centering\arraybackslash}p{2.0cm}|
        >{\centering\arraybackslash}p{1.8cm}}
    \toprule
    \multirow{2}{*}{Methods} & \multirow{2}{*}{Publication} & \multirow{2}{*}{NFE} & \multicolumn{3}{c|}{Adroit}  & \multicolumn{4}{c|}{Meta-World} & \multirow{2}{*}{\textbf{Average}} \\
     & & & Hammer & Door & Pen & Easy (21) & Medium (4) & Hard (4) & Very Hard (5) & \\  \midrule
    DP & RSS'23& 10 &16$\pm${10}&34$\pm${11}&13$\pm${2}& 50.7$\pm${6.1} & 11.0$\pm${2.5}& 5.25$\pm${2.5}& 22.0$\pm${5.0}& 35.2$\pm${5.3} \\
    Adaflow &	NeuRIPS'24 &- &45$\pm${11}& 27$\pm${6}& 18$\pm${6}& 49.4$\pm${6.8} & 12.0$\pm${5.0}& 5.75$\pm${4.0}& 24.0$\pm${4.8}& 35.6$\pm${6.1} \\
    CP & arxiv'24& 1 &45$\pm${4}& 31$\pm${10} &13$\pm${6}& 69.3$\pm${4.2} & 21.2$\pm${6.0}& 17.5$\pm${3.9}& 30.0$\pm${4.9}& 50.1$\pm${4.7} \\
    DP3 & RSS'24& 10 &100$\pm${0}&56$\pm${5}&46$\pm${10}& 87.3$\pm${2.2} & 44.5$\pm${8.7}& 32.7$\pm${7.7}& 39.4$\pm${9.0}& 68.7$\pm${4.7} \\
    Simple DP3 & RSS'24 & 10 &98$\pm${2} &40$\pm${17}&36$\pm${4}& 86.8$\pm${2.3}&42.0$\pm${6.5}&38.7$\pm${7.5}& 35.0$\pm${11.6}& 67.4$\pm${5.0} \\ 
    FlowPolicy & AAAI'25  & 1  &98$\pm${1} &61$\pm${2}& 54$\pm${4}& 84.8$\pm${2.2}& 58.2$\pm${7.9}&40.2$\pm${4.5}& 52.2$\pm${5.0}& 71.6$\pm${3.5} \\ \midrule
    MP1  & - &1  & \textbf{100}$\pm$0 & \textbf{69}$\pm$2 & \textbf{58}$\pm$5 & \textbf{88.2}$\pm$1.1 &\textbf{68.0}$\pm$3.1 & \textbf{58.1}$\pm$5.0& \textbf{67.2}$\pm$2.7& \textbf{78.9}$\pm$\textbf{2.1}
    \\ \bottomrule
    \end{tabular}
    }
    \caption{Performance of different methods on 37 Tasks. We evaluate the performance of our method on 3 Adroit and 34 Meta-World tasks with three random seeds, comparing it to SOTA methods based on Diffusion and Flow. Our method with NFE=1 outperforms the best Diffusion-based method (DP3) by 10.2\% in success rate. Compared to FlowPolicy, which uses 1-NFE sampling, our method achieves a 7.3\% higher average success rate across all tasks.}
    \label{success}
\end{table*}

\begin{table*}[t]
    \centering
    \resizebox{1.0\linewidth}{!}{
    \begin{tabular}{
        l|
        >{\centering\arraybackslash}p{1.8cm}|
        >{\centering\arraybackslash}p{1cm}|
        >{\centering\arraybackslash}p{1.6cm}
        >{\centering\arraybackslash}p{1.6cm}
        >{\centering\arraybackslash}p{1.6cm}|
        >{\centering\arraybackslash}p{1.7cm}
        >{\centering\arraybackslash}p{1.7cm}
        >{\centering\arraybackslash}p{1.7cm}
        >{\centering\arraybackslash}p{2.0cm}|
        >{\centering\arraybackslash}p{1.6cm}}
    \toprule
    \multirow{2}{*}{Methods} &\multirow{2}{*}{Publication}& \multirow{2}{*}{NFE} & \multicolumn{3}{c|}{Adroit /ms}  & \multicolumn{4}{c|}{Meta-World /ms} & \multirow{2}{*}{\textbf{Average /ms}} \\
     & & & Hammer & Door & Pen & Easy (21) & Medium (4) & Hard (4) & Very Hard (5) & \\  \midrule
    DP3 & RSS'24& 10 &129.5$\pm${13.9} & 141.3$\pm${14.8}& 145.1$\pm${12.3}& 129.3$\pm${10.7} & 134.7$\pm${11.5}& 131.9$\pm${12.4}& 138.4$\pm${10.8}& 132.2$\pm${11.2} \\
    Simple DP3 & RSS'24& 10 &103.1$\pm${11.4} &111.3$\pm${10.2}&128.2$\pm${13.1}& 91.9$\pm${8.6}&98.3$\pm${9.1}&101.3$\pm${9.7}& 103.8$\pm${10.2}& 97.0$\pm${9.2} \\ 
    FlowPolicy  & AAAI'25& 1 &15.3$\pm${1.1} &13.2$\pm${4.0}& 12.0$\pm${2.8}& 12.0$\pm${1.4}& 12.2$\pm${1.5}&13.5$\pm${1.4}& 14.5$\pm${1.6}& 12.6$\pm${1.5} \\ \midrule
    MP1 & - & 1  & \textbf{7.1}$\pm${0.2} &\textbf{7.2}$\pm${0.1}& \textbf{7.4}$\pm${0.3}& \textbf{6.7}$\pm${0.0}& \textbf{6.7}$\pm${0.1}& \textbf{6.7}$\pm${0.1}& \textbf{6.8}$\pm${0.1}& \textbf{6.8}$\pm${0.1} 
    \\ \bottomrule
    \end{tabular}
    }
    \caption{Comparison of inference times for different methods evaluated on the Meta-World and Adroit benchmark. Due to its multi-step denoising process, Diffusion-based approaches run slower than Flow-based ones. MP1 achieves SOTA inference speed across all sub-tasks, with an average latency of just 6.8 ms—nearly 2× faster than the best FlowPolicy (which relies on consistency constraints for its 1-NFE sampling )  and nearly 14× faster than Diffusion-based methods.}
    \label{inference}
\end{table*}

Crucially, $\mathcal{L}_{Disp}$, as a training-time regularizer, adds no computational overhead during inference, preserving the policy's efficiency. Given the current conditional features $\mathbf{c}$ and a random noise sample $\mathbf{A}_1 \sim \mathcal{N}(0,I)$, the MP1 generates the entire K-step action $\mathbf{A}_0$ in a single forward pass:
\begin{equation}
\mathbf{A}_{0}=\mathbf{A}_{1}-u_{\theta}^{cfg}(\mathbf{A}_{1},0,1|\mathbf{c})
\end{equation}
where $r=0, t=1$. This preserves the 1-NFE capability that is vital for real-time robotic control, while benefiting from the enhanced robustness conferred by a more structured and discriminative representational foundation.

\section{Simulation Experiments}

To demonstrate the efficacy of the MP1, we conduct fair comparative experiments on multiple datasets (e.g., Adroit \cite{adro} and Meta-World \cite{meta}), comparing our proposed method against existing approaches in terms of success rate and inference time.

\subsection{Simulation benchmark.}

In simulation, we evaluate the proposed method on three tasks from the Adroit benchmark. To further assess the algorithm’s generality and robustness across scenarios, we also select thirty-four tasks from the Meta-World benchmark, comprising twenty-one Easy tasks, four Medium tasks, four Hard tasks, and five Very Hard tasks.
\subsection{Baselines.}

In our comparison with existing SOTA methods, we include DP \cite{dp1,dp2}, AdaFlow \cite{adaflow}, and CP \cite{cp} with 2D inputs, as well as DP3 \cite{dp3}, Simple DP3 \cite{dp3}, and FlowPolicy \cite{flowpolicy} with 3D inputs. DP, DP3, and Simple DP3 all employ 10-NFE, whereas CP and FlowPolicy use 1-NFE, and AdaFlow operates with a variable NFE.

\subsection{Implementation Details.}

We generate ten expert demonstrations for training on Adroit and  Meta-World. For the point-cloud data, farthest point sampling (FPS) sampling reduces the number of points to either 512 or 1024; for the image data, resolution is downsampled to 84 × 84 pixels. 
The proposed MP1 and SOTA methods are each trained and tested using three random seeds (0, 1, and 2). For each random seed, the model is trained for 3,000 epochs on Adroit and 1,000 epochs on Meta-World, with performance evaluated every 200 epochs. From the evaluation results for each seed, the five highest success rates are selected and averaged. Finally, the overall success rate and standard deviation for the task are computed across all three seeds.
All training and testing are performed on an NVIDIA RTX 4090 GPU, with a batch size of 128, optimization uses the AdamW optimizer with a learning rate of 0.0001 (Adroit and Meta-World apply the same learning rate), an observation window of 2 steps, a history length of 4 states, and a prediction horizon of 4 steps. 
Due to varying loads, we perform inference speed tests for three seeds simultaneously on the same GPU. First, we take the three inference speed values under the stable state of the GPU and calculate their average. Then, we compute the average and standard deviation of the inference speeds for the three random seeds, which serve as the experimental results.

\begin{figure}[t]
    \centering
    \includegraphics[width=1.0\linewidth]{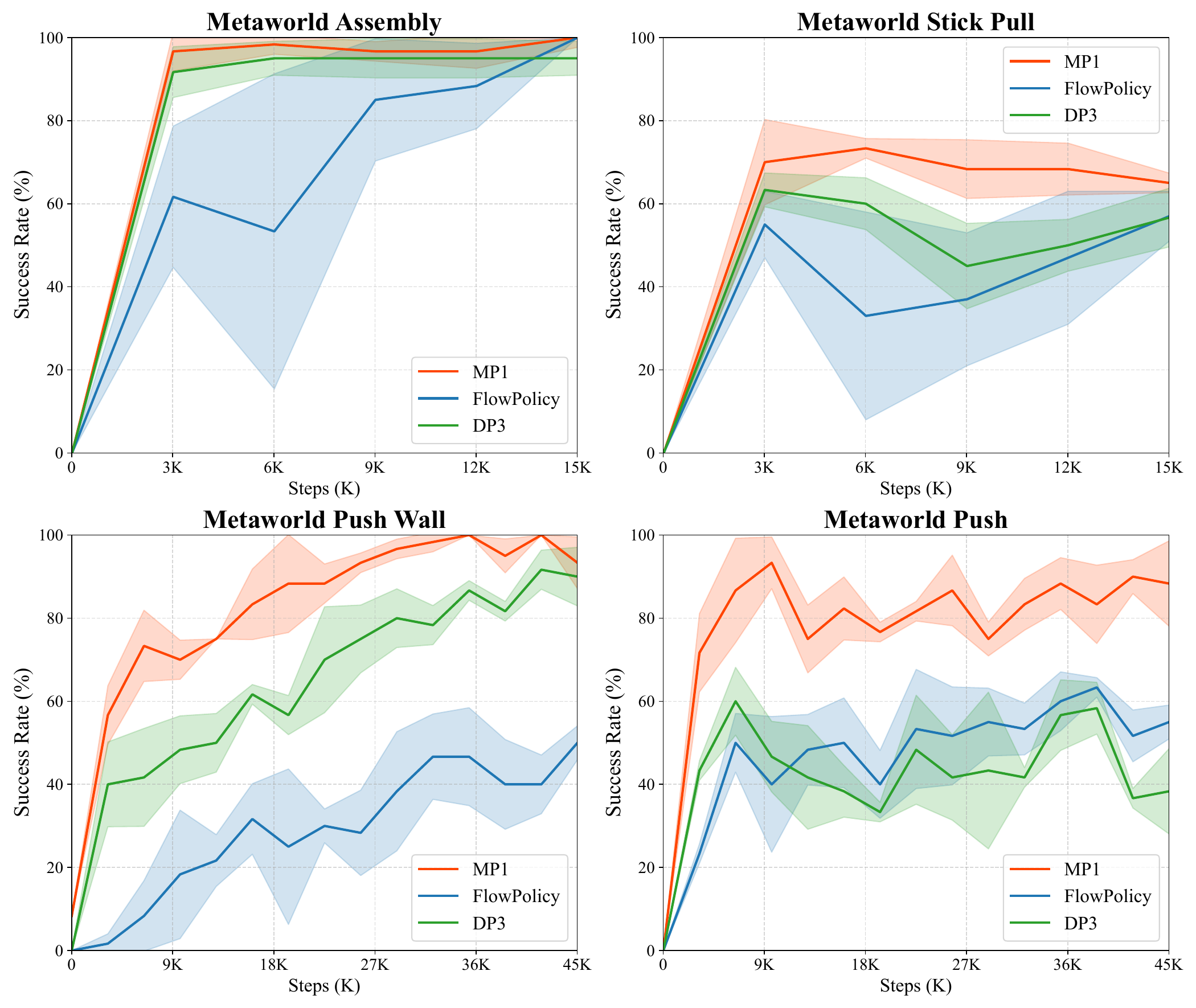}
    \caption{Success rate curves of different methods on multiple Meta-World tasks. We compare the performance of MP1, FlowPolicy, and DP3 on four tasks. The x-axis represents training steps, and the y-axis shows the success rate. Shaded areas represent the standard deviation across different random seeds. The proposed method achieves higher success rates with smaller variance.}
    \label{results}
\end{figure}
\begin{table*}[t]
    \centering
    \resizebox{\textwidth}{!}{%
    \scriptsize
    \begin{tabular}{c|c|c|c|c|c|c|c|c|c|c}
    \toprule
    \multirow{2}{*}{\makecell{\textbf{ Task /} \\ \textbf{Method}}}
    & \multicolumn{2}{c|}{\textbf{Adroit}} & \multicolumn{8}{c}{\textbf{Meta-World}} \\
    \cmidrule(lr){2-11}
    & Door & Pen & Reach & Coffee Pull & Hand Insert & Pick Place & Push & Disassemble & Stick Pull & Pick Place Wall\\
    \hline
    MP1 & 58$\pm$5 & 69$\pm$2 & 24.7$\pm$3.3 &  92.3$\pm$3.7 & 10.0$\pm$2.9 & 50.7$\pm$9.1 & 74.0$\pm$6.7 
    & 74.0$\pm$1.4 & 74.0$\pm$1.4& 64.3$\pm$1.2\\
    -$Loss_{dis}$ & 55$\pm$6 & 68$\pm$4 & 19.7$\pm$1.2 & 90.7$\pm$2.1&9.3$\pm$1.7 & 48.7$\pm$8.2 & 50.7$\pm$7.6 & 
    72.7$\pm$0.5 & 72.0$\pm$5.0&60.3$\pm$2.4\\
    \bottomrule
    \end{tabular} }
    \caption{Ablation Study on \textit{Dispersive Loss} for Adroit and Meta-World Tasks. -$Loss_{dis}$ signifies that the Dispersive Loss term has been omitted.}
    \label{ablation0}
\end{table*}
\subsection{Results of Simulation.}

Tab. \ref{success} demonstrates that the MP1 achieves SOTA performance across all sub-tasks. The overall average success rate reaches 78.9\%$\pm$2.1\%, which significantly outperforms the previous best method, FlowPolicy, at 71.6\%$\pm$3.5\%. Compared to existing approaches, our method yields a 10.2\% improvement over DP3 and a 7.3\% improvement over FlowPolicy in terms of average success rate, indicating that mean-velocity flows are better suited for the robot learning.
In the Meta-World tasks, it can be observed that the MP1 achieves a higher success rate on ``Very Hard" tasks compared to its performance on ``Hard" tasks. The proposed method demonstrates notable improvements across 13 tasks of Meta-World (Medium, Hard, and Very Hard), with success rates increased by 9.8\%, 17.9\%, and 15.0\% respectively over the FlowPolicy. On the 21 ``Easy" tasks in Meta-World, the proposed approach achieves a success rate of 88.2\%, representing a 3.4\% improvement over the FlowPolicy.
Moreover, the proposed method maintains a consistently low standard deviation on certain subtasks (for example, the standard deviation for Adroit Hammer, Door, and the Easy tasks in Meta-World ranges from only 0 to 2.0\%), and the average fluctuation in success rate is merely 2.1\%, which is lower than that of other methods. These results further indicate that the proposed approach exhibits greater stability and reliability under different random seeds.

Fig. \ref{results} shows the training success rate curves of MP1, FlowPolicy, and DP3 on four Meta-World tasks: Assembly, Stick Pull, Push Wall, and Push. As the number of training steps increases, all methods demonstrate improved success rates; however, MP1 achieves faster convergence and higher final success rates across all tasks. Furthermore, the shaded area representing the standard deviation across different random seeds is generally narrower for MP1, indicating better stability and robustness. 

Fig. \ref{scatter} and Tab. \ref{inference} summarize the inference speeds of various methods on the Adroit and Meta-world benchmarks. The results show that Diffusion-based approaches are much slower than Flow-based methods. The proposed method achieves SOTA performance in all subtasks, with approximately a 2× speedup over FlowPolicy and a 14× speedup over Simple DP3. On an NVIDIA 4090, our method attains an average inference time of 6.8 ms. Furthermore, MP1 is more stable than prior methods, unaffected by GPU load fluctuations, with a standard deviation of$\pm$0.1 ms.

Overall, Flow-based methods can achieve 1-NFE, enabling much faster inference than Diffusion-based approaches (Fig. \ref{scatter}). Our proposed MP1 achieves SOTA inference speed (6.8 ms) and performance, with a 78.9\% success rate across 37 tasks.

\section{Ablation Study}
\begin{table}[t]
    \centering
    \renewcommand{\arraystretch}{1.3}
    \resizebox{0.48\textwidth}{!}{
    \begin{tabular}{c|c|c|c|c|c|c}
    \toprule
    \multirow{2}{*}{\makecell{\textbf{ Task /} \\ \textbf{Flow Ratio}}}
    & \textbf{Adroit} & \multicolumn{4}{c|}{\textbf{Meta-World}}&
    \multirow{2}{*}{\textbf{Avg.}} \\
    \cmidrule(lr){2-6}
    & Pen & Dial Turn & Coffee Pull & Assembly & Disassemble & \\
    \hline
    0 ($r = t$)  & 53$\pm$5& 81$\pm$1 & 62$\pm$4 & 97$\pm$2& 70$\pm$3 & 72.6$\pm$3.0\\
    \hline
    0.25 & 49$\pm$6 & 89$\pm$2 & 92$\pm$0 & 99$\pm$1& 71$\pm$4 & 80.0$\pm$2.6\\
    0.50 & 58$\pm$5 & 90$\pm$2 & 92$\pm$4 & 98$\pm$1& 74$\pm$1 & 82.4$\pm$2.6\\
    0.75 & 54$\pm$4 & 90$\pm$2 & 90$\pm$5& 98$\pm$1 & 71$\pm$3
    & 80.6$\pm$3.0\\
    1.0 & 0$\pm$0  & 0$\pm$0& 12$\pm$5 & 0$\pm$0 & 0$\pm$0 & 2.4$\pm$1.0\\
    \bottomrule
    \end{tabular}
    }
    \caption{Performance of different flow ratios (when $r \neq t$) in Adroit Pen and Meta-World tasks. }
    \label{ablation1}
\end{table}
We primarily conduct ablation experiments on the dispersive loss, flow ratio, and number of demonstrations in MP1.

\subsection{Dispersive Loss.}
Tab. \ref{ablation0} compares the standard MP1 with a variant in which the Dispersive Loss is removed. Introducing the \textit{Dispersive Loss} (see the ``MP1" row) improves success rates on all ten tasks across the Adroit and Meta-World benchmarks, with average gains of roughly 4–5 percentage points. These results show that Dispersive Loss enhances policy performance in diverse manipulation scenarios.
\begin{figure}[t]
    \centering
    \includegraphics[width=1\linewidth]{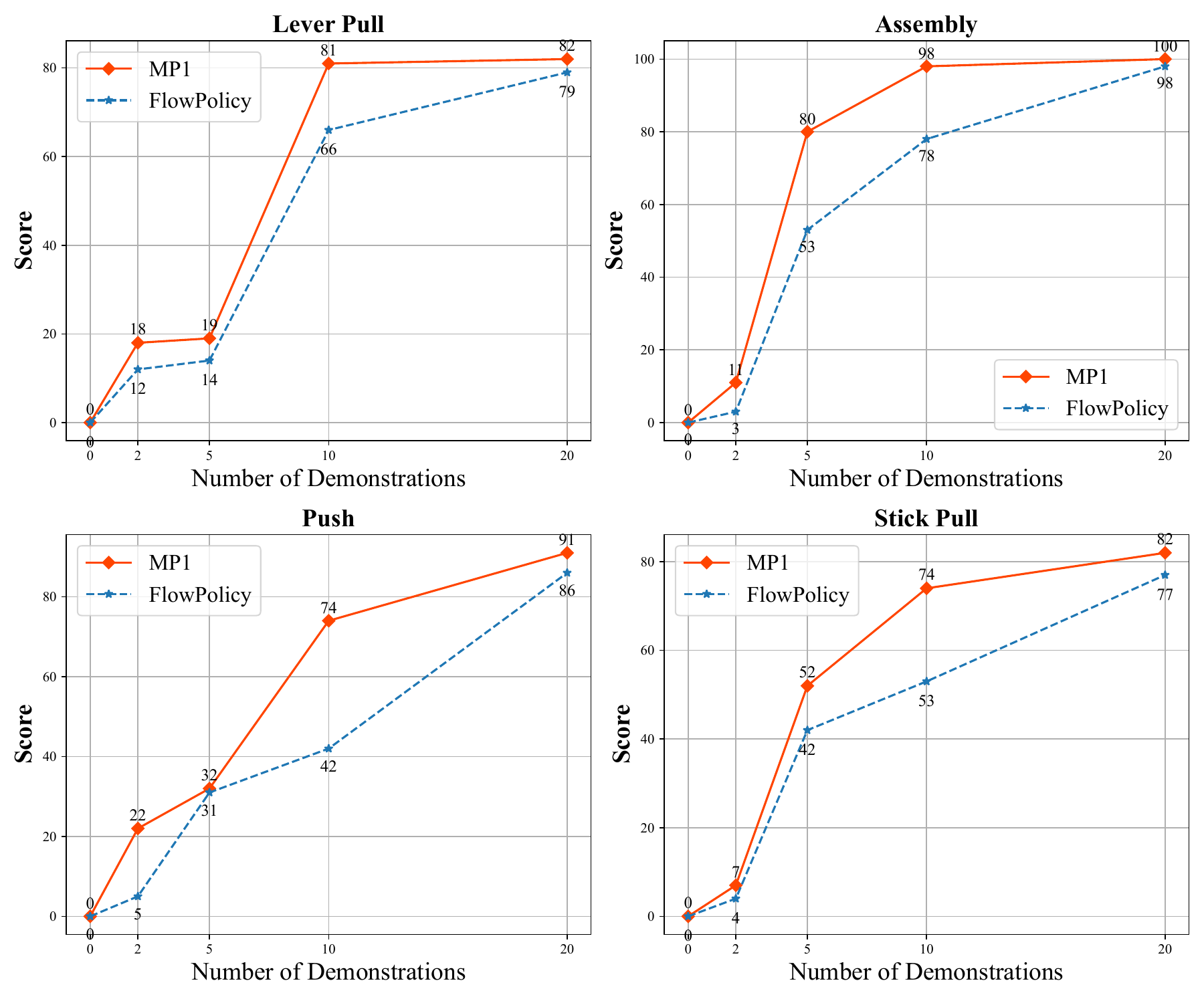}
    \caption{The effect of the number of demonstrations on different methods. As the number increases, the success rate gradually improves.}
    \label{figure:ablation}
\end{figure}

\subsection{MeanFlow Ratio.}
In \eqref{equ}, the average speed is reflected when $r\neq t$; when $r = t$ (ratio=0), it degenerates to the standard flow matching method. We extract five tasks from Adroit and Meta-World and conduct success rate tests under different flow ratios (0, 0.25, 0.50, 0.75, 1.0). As shown in Tab. \ref{ablation1}, performance declines when it reverts to Flow Matching (ratio = 0), compared to the case where $r \neq t$.
The results from Tab. \ref{ablation1} indicate that the flow ratio ($ r\neq t$) has minimal impact on task success rates for most tasks, except for extreme cases (1.0 flow ratio) where the success rate drops significantly.


\subsection{Number of Demonstrations.}
In Fig. \ref{figure:ablation}, we evaluate the impact of different numbers of demonstrations (0, 2, 5, 10, 20) on task performance. As the number of demonstrations increases, the success rate improves significantly across various tasks. 
Our method, the MP1, consistently outperforms the FlowPolicy, especially with fewer demonstrations. These results show that increasing the number of demonstrations enhances model performance for most tasks, with the proposed method excelling in few-shot learning scenarios.

\section{Real-world Experiments}

\subsection{Experimental Setup.}
\begin{figure}[t]
    \centering
    \includegraphics[width=0.9\linewidth]{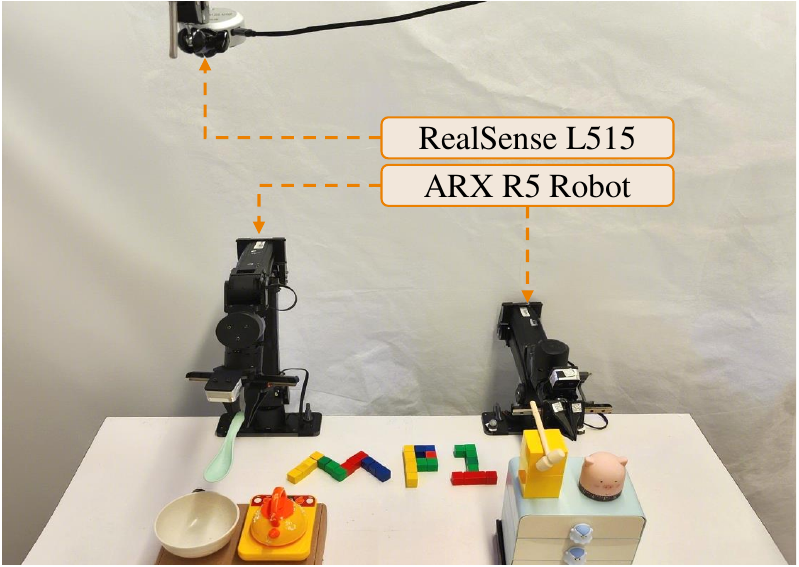}
    \caption{Real-world setup.}
    \label{setup}
\end{figure}

We conduct experiments on the ARX R5 dual-arm robot, using the same camera configuration as DP3, the RealSense L515. The experimental environment is shown in Fig. \ref{setup}.

\begin{table}[t]
    \centering
    \resizebox{\columnwidth}{!}{%
    \begin{tabular}{@{}c|c|c|c|c|c@{}}
    \toprule
    \makecell[c]{\textbf{Task /}\\\textbf{Method}}  
    & \textbf{Hammer} & \textbf{Drawer Close}&\textbf{Heat Water}&\textbf{Stack Block} &\textbf{Spoon}\\
    \midrule
    MP1 & 90\% / 18.6s & 100\% / 8.8s & 90\% / 23.4s & 80\% / 27.2s & 90\% / 22.6s \\
    FlowPolicy & 70\% / 22.3s & 90\% / 15.7s & 60\% / 31.1s & 50\% / 29.6s & 80\% / 26.7s\\
    DP3         & 70\% / 31.1s & 80\% / 20.2s & 70\% / 38.8s & 60\% / 35.1s & 70\% / 28.3s\\
    \bottomrule
    \end{tabular}
    }
    \caption{Success rates (\%) and per-task completion times (s) of different methods in real-world experiments.}
    \label{realworld_results}
\end{table}
\subsection{Real-world Tasks.}
We evaluate our approach on five real-world tasks: 1) Hammer – Grasp a hammer and strike a pig. 2) Drawer Close – Closing a drawer. 3) Heat Water – Position a kettle in a suitable location. 4) Stack Block - Stack a block. 5) Spoon - Put the spoon in the bowl.
\subsection{Real-world Experimental Results.}
Tab. \ref{realworld_results} reports the performance of different methods in real-world robotic experiments, measured by success rate (\%) and average task completion time (s). Each task was trained and evaluated using 20 human demonstrations under a unified setting of $horizon = 16$ and observation stride $n_{obs} = 2$. As observed, the mean policy achieves SOTA success rates across all five tasks while achieving the shortest average completion times, thereby substantiating its effectiveness in real-world scenarios.

\section{Conclusion}
In this paper, we address the limitations of existing Diffusion-based and Flow-based approaches by introducing MeanFlow into robot learning. The proposed MP1 eliminates the reliance on multi-step denoising, interval-wise instantaneous velocity integration, and consistency constraints. MP1 enables one-step action generation for robot manipulation by modeling the velocity field, making it suitable for real-time control. Furthermore, the introduction of Dispersive Loss helps disperse the latent embeddings of distinct states, improving generalization under low-data conditions. Experimental results demonstrate that MP1 delivers significant performance gains across various robotic tasks, confirming its effectiveness and wide applicability.


\bibliography{aaai2026}

\end{document}